\pdfoutput=1
\documentclass[11pt]{article}
\usepackage{naacl2021}
\usepackage{times}
\usepackage{latexsym}

\usepackage[T1]{fontenc}
\usepackage[utf8]{inputenc}

\usepackage{microtype}

\usepackage{url}

\usepackage{multirow}
\usepackage{tabularx}
\usepackage{ragged2e}
\usepackage{tabulary}
\usepackage{adjustbox}
\usepackage{stfloats}
\usepackage{graphicx}
\usepackage{multirow}
\usepackage{url}
\usepackage{hhline}
\usepackage{arydshln}
\usepackage{multicol}
\usepackage{booktabs}
\usepackage{svg}
\usepackage{bm}
\usepackage{amsmath}
\usepackage{amssymb}
\usepackage{hyperref}
\usepackage{fancyhdr}
\usepackage{graphicx,subcaption}
\usepackage{svg}
\usepackage{scalerel}
\usepackage{xspace}
\usepackage[boldmath]{numprint}
\usepackage{array,siunitx}
\usepackage{float}

\newcommand{\wsd}{WSD\xspace}
\newcommand{\LM}{LM}

\newcommand{\NLP}{NLP\xspace}
\newcommand{\baselinemethod}{LMGC\xspace}
\newcommand{\maskmethod}{LMGC-M\xspace}
\newcommand{\fone}{F1\xspace}

\title{Incorporating Word Sense Disambiguation in Neural Language Models}

\author{Jan Philip Wahle\textsuperscript{1}, Terry Ruas\textsuperscript{1}, Norman Meuschke\textsuperscript{1,2}, Bela Gipp\textsuperscript{1} \\
  \textsuperscript{1}University of Wuppertal, Rainer-Gruenter-Str. 21, D-42119, Wuppertal, Germany \\
  \textsuperscript{2}University of Konstanz, Universittsstrae 10, 78464, Konstanz, Germany \\
  \textsuperscript{1}\texttt{last@uni-wuppertal.de} \\
  \textsuperscript{2}\texttt{first.last@uni-konstanz.de}\\}

\date{23.11.2020}

\begin{document}
% \raggedbottom
% \setlength{\parskip}{0pt}
\maketitle

\begin{abstract}
We present two supervised (pre-)training methods that incorporate gloss definitions from lexical resources to leverage Word Sense Disambiguation (\wsd) capabilities in neural language models. Our training focuses on \wsd but keeps its capabilities when transferred to other tasks while adding almost no additional parameters. We evaluate our technique on 15 downstream tasks, e.g., sentence pair classification and \wsd. We show that our methods exceed comparable state-of-the-art techniques on the SemEval and Senseval datasets as well as increase the performance of its baseline on the GLUE benchmark.
\end{abstract}

%%%%%%%%%%%%%%%%%%%%%%%%%%%%%%%%%%%%%%%%%%%%%%%%%%%%%%%%%%%%%%%%%%%%%%%%%%%%%%%%%%%%%%
% INTRODUCTION
%%%%%%%%%%%%%%%%%%%%%%%%%%%%%%%%%%%%%%%%%%%%%%%%%%%%%%%%%%%%%%%%%%%%%%%%%%%%%%%%%%%%%%
\section{Introduction}

% Expain WSD
% Explain why transformers are important
% Explain the gap that everryone only uses BERT
% Explain why we should care to use WSD in pre-training

\wsd tries to determine the meaning of words given a context and is arguably one of the oldest challenges in natural language processing (NLP)~\cite{Weaver1955,Navigli09}. In knowledge-based methods~\cite{CamachoColladosPN15}, lexical knowledge databases (LKB) (e.g., WordNet~\cite{Miller95a, Fellbaum98a}) illustrate the relation between words and their meaning. Supervised techniques~\cite{PasiniN20} rely on annotated data to perform disambiguation while unsupervised ones~\cite{ChaplotS18} explore other aspects, e.g., larger contexts, topic modeling.

Recently, supervised methods~\cite{HuangSQH19, BevilacquaN20} rely on word representations from BERT~\cite{DevlinCLT19}, although advances in bidirectional transformers have been proposed~\cite{YangDYC19, ClarkLLM20}. We compare these novel models and validate which ones are most suitable for the \wsd task. Thus, we define an end-to-end approach capable of being applied to any language model (\LM).

Further, pre-trained word representations have become crucial for \LM s and almost any \NLP task \cite{MikolovCCD13,RadfordWCL18}.
LMs are trained on large unlabeled corpora and often ignore relevant information of word senses in LKB (e.g., gloss\footnote{Brief definition of a synonym set (synset)~\cite{Miller95a}.}). As our experiments support, there is a positive correlation between the LM's ability to disambiguate words and NLP tasks performance. %We show \wsd can be incorporated in the pre-training step and leverage the semantic representation of LM to other tasks (e.g., text-classification, text-similarity).

We propose a set of general supervised methods that integrate WordNet knowledge for \wsd in \LM~during the pre-training phase and validate the improved semantic representations on other tasks (e.g., text-similarity). Our technique surpasses comparable methods in WSD by 0.5\% \fone and improves language understanding in several tasks by 1.1\% on average. The repository for all experiments is publicly available\footnote{\label{github}\url{https://github.com/jpelhaW/incorporating_wsd_into_nlm}}.

% Closing statement - work towards methods
%%%%%%%%%%%%%%%%%%%%%%%%%%%%%%%%%%%%%%%%%%%%%%%%%%%%%%%%%%%%%%%%%%%%%%%%%%%%%%%%%%%%%%
% RELATED WORK
%%%%%%%%%%%%%%%%%%%%%%%%%%%%%%%%%%%%%%%%%%%%%%%%%%%%%%%%%%%%%%%%%%%%%%%%%%%%%%%%%%%%%%

\section{Related Work}
\label{sec:rw}
The same way word2vec~\cite{MikolovSCC13} inspired many models in NLP~\cite{BojanowskiGJM17, RuasFGd20}, BERT~\cite{DevlinCLT19} echoed in the literature with recent models as well~\cite{YangDYC19, ClarkLLM20}. These novel models achieve higher performance in several NLP tasks but are mostly neglected in the \wsd domain~\cite{WiedemannRCB19} with a few exceptions~\cite{LoureiroRPC20}.

Based on the Transformer~\cite{VaswaniSPU17} architecture, BERT~\cite{DevlinCLT19} proposes two pre-training tasks to capture general aspects of the language, i.e., \textit{Masked Language Model} (MLM) and \textit{Next Sentence Prediction} (NSP). AlBERT~\cite{LanCGG19}, DistilBERT~\cite{SanhDCW19}, and RoBERTa~\cite{LiuOGD19} either boost BERT's performance through parameter adjustments, increased training volume, or make it more efficient. XLNet~\cite{YangDYC19} focuses on improving the training objective, while ELECTRA~\cite{ClarkLLM20} and BART~\cite{LewisLGG19} propose a discriminative denoising method to distinguish real and plausible artificial generated input tokens.

Directly related to our work, GlossBERT~\cite{HuangSQH19} uses WordNet's glosses to fine-tune BERT in the WSD task. GlossBERT classifies a marked word in a sentence into one of its possible definitions. KnowBERT (KBERT)~\cite{PetersNLS19} incorporates LKB into BERT with a knowledge attention and recontextualization mechanism. \newcite{PetersNLS19} best-performing model, i.e., KBERT-W+W, surpasses BERT$_{BASE}$ at the cost of $\approx$ 400M parameters and 32\% more training time. Our methods do not require embeddings adjustments from the LKB or use word-piece attention, resulting in a cheaper alternative. Even though recent contributions in \wsd such as LMMS~\cite{LoureiroJ19}, BEM~\cite{BlevinsZ20}, GLU~\cite{HadiwinotoNG19}, and EWISER~\cite{BevilacquaN20} enhance the semantic representation via context or external knowledge, they do not explore their generalization to other NLP tasks.

%. Yet, \methodName, often outperforms preceding techniques in WSD and obtains the highest score in 7 out of 9 GLUE tasks.

%%%%%%%%%%%%%%%%%%%%%%%%%%%%%%%%%%%%%%%%%%%%%%%%%%%%%%%%%%%%%%%%%%%%%%%%%%%%%%%%%%%%%%
% METHODS & CONTRIBUTION
%%%%%%%%%%%%%%%%%%%%%%%%%%%%%%%%%%%%%%%%%%%%%%%%%%%%%%%%%%%%%%%%%%%%%%%%%%%%%%%%%%%%%%
\section{Methods} \label{sec:methods}

Current methods~\cite{HuangSQH19, DuQS19, PetersNLS19, LevineLDP19} modify the WSD task into a text classification problem, leveraging BERT's semantic information through WordNet's resources. Although BERT is a strong baseline, studies show the model does not converge to its full capacity, and its training scheme still presents opportunities for development~\cite{LiuOGD19, YangDYC19}.

We define a general method to perform WSD in arbitrary \LM s and discuss possible architectural alternatives for its improvements (Section~\ref{sec:lmgc}). We assume WSD is a suitable task to complement MLM as we often find polysemous words in natural text. We introduce a second variation in our method (Section~\ref{sec:lmgc-m}) that keeps previous \LM~capabilities while improving polysemy understanding.

\subsection{Language Model Gloss Classification}
\label{sec:lmgc}

With Language Model Gloss Classification (\baselinemethod), we propose a general end-to-end WSD approach to classify ambiguous words from sentences into one of WordNet's glosses. This approach allows us to evaluate different \LM s at WSD.

\baselinemethod builds on the final representations of its underlying transformer with a classification approach closely related to~\cite{HuangSQH19}. Each input sequence starts with an aggregate token (e.g., the ``[CLS]'' token in BERT), i.e., an annotated sentence containing the ambiguous word, followed by a candidate gloss definition from a lexical resource, such as WordNet, for that specific word.

Sentence and gloss are concatenated with a separator token and pre-processed with the respective model's tokenizer.
We modify the input sequence with two supervision signals: (1) highlighting the ambiguous tokens with two special tokens and (2) adding the polysemous word before the gloss.

Considering \newcite{DuQS19} and \newcite{HuangSQH19} findings, we apply a linear layer to the aggregate representation of the sequence to perform classification rather than using token embeddings. In contrast, we suggest modifying the prediction step from sequential binary classification to a parallel multi-classification construct, similar to \newcite{KagebackS16}. Therefore, we stack the $k$ candidate sentence-gloss pairs at the batch dimension and classify them using softmax. 

We retrieve WordNet's gloss definitions of polysemous words corresponding to the annotated synsets to create sentence-gloss inputs. %Table~\ref{table:dataset_sem} shows the imbalanced labeled examples for the Semcor datasets. 
To accelerate training time by approximately a factor of three, compared to~\cite{HuangSQH19}, we reduced the sequence length of all models from 512 to 160\footnote{99.8\% of the data set can be represented with 160 tokens; we truncate remaining sequences to this limit.} as the computational cost of transformers is quadratic concerning the sequence length.

\subsection{\baselinemethod with Masked Language Modeling}
\label{sec:lmgc-m}

% \baselinemethod is a general approach to perform WSD with a variety of auto encoding neural LMs to identify the most efficient one.

%To find the most effective transformer architecture performing WSD, we proposed \baselinemethod.
Comparable \wsd systems~\cite{HuangSQH19, BevilacquaN20, BlevinsZ20} and \baselinemethod focus on improving the performance in WSD rather than leveraging it with lexical resources for language understanding.
We assume the transfer learning between LM and WSD increases the likelihood of grasping polysemous words in co-related tasks.
Thus, we employ \baselinemethod as an additional supervised training objective into MLM (\maskmethod) to incorporate lexical knowledge into our pre-training.

%We assume the transfer learning between LM and WSD increases the likelihood of grasping polysemous words in co-related tasks.

% In Appendix~\hyperref[appx:A]{A}, we discuss the correlation between WSD and other NLP tasks concerning the number of polysemous words.

\maskmethod performs a forward pass with annotated examples from our corpus with words masked at a certain probability. Moreover, \maskmethod uses \baselinemethod as a second objective, similar to NSP in BERT. To prevent underfitting, due to task difficulty, we only mask words in the context of the polysemous word. %The procedure is exemplified in Figure~\ref{fig:methods}
Before inference, we fine-tune \baselinemethod without masks.

% \begin{figure}[t]
%     \centering
%     \includegraphics[scale=0.70]{figures/BGP-M.pdf}
%     %\caption{\maskmethod}
%     %\label{fig:BGP-M}
% \caption{\maskmethod forwards context-gloss pairs and its masked versions through BERT to perform a weakly supervised softmax classification (\baselinemethod) and MLM.}
% \label{fig:methods}
% \end{figure}

%%%%%%%%%%%%%%%%%%%%%%%%%%%%%%%%%%%%%%%%%%%%%%%%%%%%%%%%%%%%%%%%%%%%%%%%%%%%%%%%%%%%%%
% EXPERIMENTS - SETUP & DATA
%%%%%%%%%%%%%%%%%%%%%%%%%%%%%%%%%%%%%%%%%%%%%%%%%%%%%%%%%%%%%%%%%%%%%%%%%%%%%%%%%%%%%%

\section{Experiments} \label{sec:experiments}

We evaluate our proposed methods in two benchmarks, namely SemCor (3.0)~\cite{MillerLTB93, RaganatoCN17} and GLUE~\cite{WangSMH19}. The English all-words WSD benchmark SemCor, detailed in Table~\ref{table:dataset_sem}, is popular in the \wsd literature~\cite{HuangSQH19, PetersNLS19} and one of the largest manually annotated datasets with approximately 226k word sense annotations from WordNet~\cite{Miller95a}. GLUE~\cite{WangSMH19} is a collection of nine language understanding tasks widely used~\cite{DevlinCLT19,LanCGG19} to validate the generalization of LMs in different linguistic phenomena. All GLUE tasks are single sentence or sentence pair classification, except STS-B, which is a regression task. 

\begin{table}[h]
\centering
\resizebox{0.48\textwidth}{!}{
    \begin{tabular}{lrrrrrrrrr} \toprule
    \multirow{2}{*}{\textbf{Dataset}} & \multicolumn{5}{c}{\textbf{POS Tags}} & \multicolumn{2}{c}{\textbf{Class dist.}} \\
    \cmidrule(lr){2-6}
    \cmidrule(lr){7-8}
    {} & Noun   & Verb   & Adj. & Adv. & Total & Pos. & Neg. \\ \midrule
    SemCor  & 87k & 88.3k & 31.7k   & 18.9k & 226k & 226.5k & 1.79m \\
    SE2     & 1k & 517 & 445 & 254 & 2.3k & 2.4k & 14.2k \\
    SE3     & 900 & 588 & 350 & 12 & 1.8k & 1.8k & 15.3k \\
    SE7    & 159 & 296 & 0 & 0  & 455 & 459 & 4.5k \\
    SE13    & 1.6k & 0 & 0 & 0 & 1.6k & 1.6k & 9.7k \\
    SE15    & 531     & 251     & 160       & 80 & 1k & 1.2k & 6.5k \\        \bottomrule 
\end{tabular}}
\caption{SemCor training corpus details: general statistics (left) and class distribution for \baselinemethod (right).} %{\cite{MillerLTB93}} and {\newcite{RaganatoCN17}} evaluation  framework. }
\label{table:dataset_sem}
\end{table}

% \begin{table}[h]
% \centering
% \resizebox{0.48\textwidth}{!}{
%     \begin{tabular}{lrrrrrrrrr} \toprule
%     \multirow{2}{*}{\textbf{Dataset}} & \multicolumn{5}{c}{\textbf{POS Tags}} & \multicolumn{2}{c}{\textbf{Class dist.}} \\
%     \cmidrule(lr){2-6}
%     \cmidrule(lr){7-8}
%     {} & Noun   & Verb   & Adjective & Adverb & Total & Pos. Labels & Neg. Labels \\ \midrule
%     SemCor  & 87\,002 & 88\,334 & 31\,753   & 18\,947 & 226\,036 & 226\,563 & 1\,794\,040 \\
%     SE2     & 1\,066 & 517 & 445 & 254 & 2\,282 & 2\,389 & 14\,247 \\
%     SE3     & 900 & 588 & 350 & 12 & 1\,850 & 1\,888 & 15\,346 \\
%     SE07    & 159 & 296 & 0 & 0  & 455 & 459 & 4\,527 \\
%     SE13    & 1\,644 & 0 & 0 & 0 & 1\,644 & 1\,656        & 9\,732        \\
%     SE15    & 531     & 251     & 160       & 80 & 1\,022 & 1\,219        & 6\,558 \\        \bottomrule 
% \end{tabular}}
% \caption{SemCor 3.0 training corpus details: POS general statistics (left) and class distribution (right).} %{\cite{MillerLTB93}} and {\newcite{RaganatoCN17}} evaluation  framework. }
% \label{table:dataset_sem}
% \end{table}

\subsection{Setup}
All models were initialized using the base configuration of its underlying transformer (e.g., BERT$_{BASE}$, L=12, H=768, A=12). Both of our methods have $2*H+2$ more parameters compared to their baseline (e.g., \baselinemethod (BERT) has $\approx$ 110M parameters). We increased the hidden dropout probability to $0.2$ as we observed overfitting for most models. Further, we explicitly treated the class imbalance of positive and negative examples (Table~\ref{table:dataset_sem}) in \baselinemethod with focal loss~\cite{LinGGH17} ($\gamma=2$, $\alpha=0.25$).
Following~\newcite{DevlinCLT19}, we used a batch size of 32 sequences, the AdamW optimizer ($\alpha$ = 2e-5), trained three epochs, and choose the best model by validation loss. We applied the same hyperparameter configuration for all models used in both SemCor and GLUE benchmarks. The training was performed on 1 NVIDIA Tesla V100 GPU for $\approx$ 3 hours per epoch.

%To accelerate training time by a factor of three compared, to~\cite{HuangSQH19}, we reduced the sequence length of all models from 512 to 160 as the validation dataset can be entirely represented with 160 tokens\footnote{99.8\% of the training set can be represented with 160 tokens; remaining sequences are truncated to this limit.} for all tokenizers tested and compute in transformers is quadratic in relation to sequence length. We applied the same hyperparameter configuration for both SemCor and GLUE experiments.

For all GLUE tasks, except for STS-B, we transformed the aggregate embedding into a classification vector applying a new weight matrix $W \in \mathbb{R}^{K \times H}$; where $K$ is the number of labels. For STS-B, we applied a new weight matrix $V \in \mathbb{R}^{1 \times H}$ transforming the aggregate into a single value. 

\subsection{Results \& Discussion}

Table~\ref{table:model_comparison} reports the results of applying \baselinemethod to different transformer models. Our rationale for choosing models was two-fold. First, we explore models closely related or based on BERT, either by improving it through additional training time and data (RoBERTa), or compressing the architecture with minimal performance loss (DistilBERT, AlBERT). Second, models that significantly change the training objective (XLNet), or employ a discriminative learning approach (ELECTRA, BART).
In Table~\ref{table:semeval}, we compare our techniques to other contributions in WSD. All results of SemCor are reported according to~\newcite{RaganatoCN17}.

\begin{table}[t]
\centering
\resizebox{0.5\textwidth}{!}{ 
    \begin{tabular}{lcccccc} 
    \toprule
    \textbf{System} & \textbf{SE7} & \textbf{SE2} & \textbf{SE3} & \textbf{SE13} & \textbf{SE15} & \textbf{All} \\ \midrule
    BERT \citeyear{DevlinCLT19} & 71.9 & 77.8 & 74.6 & 76.5 & 79.7 & 76.6 \\\hdashline
    RoBERTa \citeyear{LiuOGD19} & 69.2 & 77.5 & 73.8 & 77.2 & 79.7 & 76.3 \\
    DistilBERT \citeyear{SanhDCW19} & 66.2 & 74.9 & 70.7 & 74.6 & 77.1 & 73.5  \\
    AlBERT (\citeyear{LanCGG19}) & 71.4 & 75.9 & 73.9 & 76.8 & 78.7 & 75.7 \\
    BART (\citeyear{LewisLGG19}) & 67.2 & 77.6 & 73.1 & 77.5 & 79.7 & 76.1 \\
    XLNet (\citeyear{YangDYC19}) & \textbf{72.5} & \textbf{78.5} & \textbf{75.6} & \textbf{79.1} & \textbf{80.1} & \textbf{77.2} \\
    ELECTRA (\citeyear{ClarkLLM20}) & 62.0 & 71.5 & 67.0 & 73.9 & 76.0 & 70.9 \\
    \bottomrule
    \end{tabular}
   }
\caption{SemCor test results of \baselinemethod for base transformer models. \textbf{Bold} font indicates the best results.}
\label{table:model_comparison}
\end{table}

\begin{table}[t]
\centering
 \resizebox{0.5\textwidth}{!}{ %TR: this is the commando I mentioned to you. I also changed the hlines to top/mid/bottom rule
    \begin{tabular}{lcccccc} \toprule
    \textbf{System} & \textbf{SE7} & \textbf{SE2} & \textbf{SE3} & \textbf{SE13} & \textbf{SE15} & \textbf{All} \\ \midrule
    GAS (\citeyear{LouLHX2018b})    &  - & 72.2 & 70.5 & 67.2 & 72.6 & 70.6 \\
    CAN (\citeyear{LuoLHX18})                   &  - & 72.2 & 70.2 & 69.1 & 72.2 & 70.9 \\
    HCAN (\citeyear{LuoLHX18})                    & - & 72.8 & 70.3 & 68.5 & 72.8 & 71.1\\
    LMMS$_{BERT}$ (\citeyear{LoureiroJ19})                              & 68.1 & 76.3 & 75.6 & 75.1 & 77.0 & 75.4 \\
    GLU (\citeyear{HadiwinotoNG19})        & 68.1 & 75.5 & 73.6 & 71.1 & 76.2 & 74.1 \\    
    GlossBERT (\citeyear{HuangSQH19})        & 72.5 & 77.7 & 75.2 & 76.1 & 80.4 & 77.0 \\ 
    BERT$_{WSD}$  (\citeyear{DuQS19})                          & - & 76.4 & 74.9 & 76.3 & 78.3 & 76.3 \\
    KBERT-W+W (\citeyear{PetersNLS19})           & - & - & - & - & - & 75.1 \\ \hdashline
    \baselinemethod (BERT) & 71.9 & 77.8 & 74.6 & 76.5 & 79.7 & 76.6 \\
    \maskmethod (BERT) & 72.9 & 78.2 & 75.5 & 76.3 & 79.5 & 77.0 \\
    \baselinemethod (XLNet) & 72.5 & 78.5 & 75.6 & 79.1 & 80.1 & 77.2 \\
    \maskmethod (XLNet) & \textbf{73.0} & \textbf{79.1} & \textbf{75.9} & \textbf{79.0} & \textbf{80.3} & \textbf{77.5} \\
    \bottomrule
    \end{tabular}
    }
\caption{SemCor test results compared to state-of-the-art techniques. \textbf{Bold} font indicates the best results.}
\label{table:semeval}
\end{table}

\begin{table*}[!ht]
\centering
 \resizebox{0.8\textwidth}{!}{
    \begin{tabular}{lccccccccc} \toprule
    \multirow{3}{*}{\textbf{System}} & \multicolumn{2}{c}{\textbf{Classification}} & \multicolumn{3}{c}{\textbf{Semantic Similarity}} & \multicolumn{3}{c}{\textbf{Natural Language Inference}} & \multicolumn{1}{c}{\textbf{Average}} \\
    \cmidrule(lr){2-3}
    \cmidrule(lr){4-6}
    \cmidrule(lr){7-9}
    \cmidrule(lr){10-10}
    {} & CoLA & SST-2 & MRPC & STS-B & QQP & MNLI & QNLI & RTE & - \\
    {} & (mc) & (acc) & (F1) & (sc) & (acc) & m/mm(acc) & (acc) & (acc) & - \\ \midrule
    BERT$_{BASE}$ & 52.1 & 93.5 & \textbf{88.9} & 85.8 & 89.3 & 84.6/83.4 & \textbf{90.5} & 66.4 & 81.4  \\
    GlossBERT & 32.8 & 90.4 & 75.2 & 90.4 & 68.5 & 81.3/80 & 83.6 & 47.3 & 70.7 \\ \hdashline
    \baselinemethod (BERT) & 31.1 & 89.2 & 81.9 & 89.2 & 87.4 & 81.4/80.3 & 85.4 & 60.2 & 74.5 \\
    \maskmethod (BERT) & \textbf{55.0} & \textbf{94.2} & 87.1 & \textbf{88.1} & \textbf{90.8} & \textbf{85.3/84.2} & 90.1 & \textbf{69.7} & \textbf{82.5}\\
    \bottomrule
    \end{tabular}
}
\caption{GLUE test results. As in BERT, we exclude the problematic WNLI set. We report F1-score for MRPC, Spearman correlations (sc) for STS-B, Matthews correlations (mc) for CoLA, and accuracy (acc) for the other tasks (with matched/mismatched accuracy for MNLI). \textbf{Bold} font indicates the best results.}
\label{table:glue}
\end{table*}

RoBERTa shows inferior \fone when compared to BERT although it uses more data and training time. As expected, DistilBERT and AlBERT perform worse than BERT, but AlBERT keeps reasonable performance with only $\approx$ 10\% of BERT's parameters. ELECTRA and BART results show their denoising approach is not suitable for our WSD setup. Besides, BART presents similar performance to BERT, but with ~26\% more parameters. XLNet constantly performs better than BERT on all evaluation sets with no additional parameters, justifying its choice for our models' variations.

We see an overall improvement when comparing \baselinemethod to the other approaches in Table~\ref{table:semeval}. \baselinemethod (BERT) generally outperforms the baseline BERT$_{WSD}$ approach, and KBERT-W+W which has four times the number of parameters. We show by using an optimal transformer (XLNet) and adjustments in the training procedure, we can outperform GlossBERT in all test sets. We exclude EWISER~\cite{BevilacquaN20} which explores additional knowledge other than gloss definition (e.g, knowledge graph). We leave for future work the investigation of BEM~\cite{BlevinsZ20}, a recently published bi-encoder architecture with two encoders (i.e., context and gloss) that are learned simultaneously.

%We show by using the best-suited transformer (XLNet) and adjustments in the training procedure, we can outperform GlossBERT in all test sets.

\maskmethod often outperforms \baselinemethod, which we assume is due to the similarity to discriminated fine-tuning~\cite{HowardR18}. We combine \baselinemethod and MLM in one pass, achieving higher accuracy in WSD and improving generalization. Considering large models, preliminary experiments\textsuperscript{\ref{github}} showed a difference of 0.08\% in \fone between BERT$_{BASE}$ and BERT$_{LARGE}$ for the SemCor datasets which is in line with \newcite{BlevinsZ20}. Thus, we judged the base configuration sufficient for our experiments.

To show that WSD training allows language models to achieve higher generalization, we fine-tune the weights from our approaches in the GLUE~\cite{WangSMH19} datasets. Our results in Tables~\ref{table:semeval} and~\ref{table:glue} show \maskmethod outperforms the state-of-the-art in the WSD task and successfully transfer the acquired knowledge to general language understanding datasets. We exclude XLNet from the comparison to show that the additional performance can be contributed mainly to our method; not to the improvement of XLNet over BERT.
The number of polysemous words in the GLUE benchmark is high in general, supporting the training design of our method. We provide more details about polysemy in GLUE in our repository\textsuperscript{\ref{github}}.

We evaluated our proposed methods against the best performing model in WSD (Table~\ref{table:semeval}) on the GLUE datasets (Table~\ref{table:glue}). Comparing \maskmethod with the official BERT$_{BASE}$ model, we achieve a 1.1\% increase in performance on average. In this work, we did not compare \maskmethod to the other WSD methods performing worse than \newcite{HuangSQH19} in the WSD task (Table~\ref{table:semeval}) because of computational requirements (i.e., KBERT-W+W is 32\% slower). Unsurprisingly, \baselinemethod and GlossBERT perform well in WSD, but cannot maintain performance on other GLUE tasks. \maskmethod outperforms the underlying baseline (BERT) on most tasks and is comparable to the others. Therefore, incorporating MLM to our WSD architecture leverages \baselinemethod  semantic representation and improves its natural language understanding capabilities.

% Still, \baselinemethod performs better than GlossBERT because of weak supervision without token insertion and a less biased loss function to negative examples.

%%%%%%%%%%%%%%%%%%%%%%%%%%%%%%%%%%%%%%%%%%%%%%%%%%%%%%%%%%%%%%%%%%%%%%%%%%%%%%%%%%%%%%
% CONCLUSION AND FUTURE WORK
%%%%%%%%%%%%%%%%%%%%%%%%%%%%%%%%%%%%%%%%%%%%%%%%%%%%%%%%%%%%%%%%%%%%%%%%%%%%%%%%%%%%%%

\section{Conclusions and Future Work}
In this paper, we proposed a set of methods (\baselinemethod, \maskmethod) that allows for (pre-)training WSD, which is essential for many NLP tasks (e.g., text-similarity). Our techniques perform WSD combining neural language models with lexical resources from WordNet. We exceeded state-of-the-art of WSD methods (+0.5\%) and improved the performance over BERT in general language understanding tasks (+1.1\%). Future work will include testing generalization on the WiC~\cite{PilehvarC19}, and SuperGLUE~\cite{WangPNS19} datasets. Besides, we want to test disciminative fine-tuning against our parallel approach ~\cite{HowardR18}, and perform an ablation study to investigate which components of our methods lead to the most benefits. We also leave for future work to incorporate knowledge from other sources (e.g., Wikidata, Wikipedia).  

%\section*{Impact Statement}
%Pre-trained language models require large corpora and enormous compute budgets to achieve high performance \cite{Kaplan20}. 
%We showed that by choosing a suitable model for the task at hand, and with few additional gradient steps, we can make use of external knowledge to improve the model more efficiently. By questioning model choices (e.g., sequence length), we were able to reduce training time up to three times compared to state-of-the-art methods. Practical applications can benefit from these efficient systems rather than scaling data and compute.

%%%%%%%%%%%%%%%%%%%%%%%%%%%%%%%%%%%%%%%%%%%%%%%%%%%%%%%%%%%%%%
% ACKNOWLEDGEMENTS
%%%%%%%%%%%%%%%%%%%%%%%%%%%%%%%%%%%%%%%%%%%%%%%%%%%%%%%%%%%%%%%
% to be included only for the camera ready version
%\section*{Acknowledgements}
%The acknowledgements should go immediately before the references.  Do not number the acknowledgements section. Do not include this section when submitting your paper for review.

%%%%%%%%%%%%%%%%%%%%%%%%%%%%%%%%%%%%%%%%%%%%%%%%%%%%%%%%%%%%%%
% BIBLIOGRAPHY
\bibliography{_references.bib}
\bibliographystyle{acl_natbib}

%%%%%%%%%%%%%%%%%%%%%%%%%%%%%%%%%%%%%%%%%%%%%%%%%%%%%%%%%%%%%%
% APPENDIX
%%%%%%%%%%%%%%%%%%%%%%%%%%%%%%%%%%%%%%%%%%%%%%%%%%%%%%%%%%%%%%%
% {\bf Appendices}: Appendices, if any, directly follow the text and the
% references (but see above).  Letter them in sequence and provide an
% informative title: {\bf Appendix A. Title of Appendix}.

\end{document}